\documentclass[sigconf]{acmart}

\setcopyright{none}
\renewcommand\footnotetextcopyrightpermission[1]{} 
\AtBeginDocument{%
  }

\copyrightyear{2018}
\acmYear{2018}
\acmDOI{XXXXXXX.XXXXXXX}
\acmConference[Conference acronym 'XX]{Make sure to enter the correct
  conference title from your rights confirmation emai}{June 03--05,
  2018}{Woodstock, NY}
\acmISBN{978-1-4503-XXXX-X/18/06}

\usepackage{xspace}
\usepackage{subfigure}
\usepackage{caption}
\usepackage{afterpage}
\usepackage{graphicx}
\usepackage{float} 
\usepackage{subcaption}
\usepackage{pifont}

\usepackage{circledsteps}

\newcommand{\hi}[1]{\vspace{.25em} \noindent {\bf #1}\xspace}
\newcommand{\bfit}[1]{\textbf{\textit{#1}}}
\newcommand{\oursys}{\texttt{FeatInsight}\xspace}

\newcommand{\blue}[1]{\textcolor{blue}{#1}}

\setcopyright{none}
\renewcommand\footnotetextcopyrightpermission[1]{} 
\begin{document}

\title{ST-Raptor: An Agentic System for Semi-Structured Table QA}


\author{Jinxiu Qu}
\affiliation{\institution{{Shanghai Jiao Tong Univ.}}\country{}}
\email{{afuloowa@sjtu.edu.cn}}

\author{Zirui Tang}
\affiliation{\institution{{Shanghai Jiao Tong Univ.}}\country{}}
\email{{raymondtangzirui@163.com}}

\author{Hongzhang Huang}
\affiliation{\institution{{Shanghai Jiao Tong Univ.}}\country{}}
\email{{haohaizi@sjtu.edu.cn}}

\author{Boyu Niu}
\affiliation{\institution{{Shanghai Jiao Tong Univ.}}\country{}}
\email{nby2005@sjtu.edu.cn}

\author{Wei Zhou}
\affiliation{\institution{{Shanghai Jiao Tong Univ.}}\country{}}
\email{{weizhoudb@sjtu.edu.cn}}

\author{Jiannan Wang}
\affiliation{
\institution{Tsinghua University}\country{}
}
\email{jnwang@tsinghua.edu.cn}

\author{Yitong Song}
\affiliation{
\institution{Hong Kong Baptist Univ.}\country{}
}
\email{yitong_song@hkbu.edu.hk}

\author{Guoliang Li}
\affiliation{\institution{{Tsinghua University}}\country{}}
\email{{liguoliang@tsinghua.edu.cn}}

\author{Xuanhe Zhou}
\affiliation{\institution{{Shanghai Jiao Tong Univ.}}\country{}}
\email{{zhouxuanhe@sjtu.edu.cn}}

\author{Fan Wu}
\affiliation{\institution{{Shanghai Jiao Tong Univ.}}\country{}}
\email{{fwu@cs.sjtu.edu.cn}}

\renewcommand{\shortauthors}{Jinxiu Qu et al.}

\begin{abstract}


Semi-structured table question answering (QA) is a challenging task that requires (1) precise extraction of cell contents and positions and (2) accurate recovery of key implicit logical structures, hierarchical relationships, and semantic associations encoded in table layouts. In practice, such tables are often interpreted manually by human experts, which is labor-intensive and time-consuming. However, automating this process remains difficult. Existing Text-to-SQL methods typically require converting semi-structured tables into structured formats, inevitably leading to information loss, while approaches like Text-to-Code and multimodal LLM-based QA struggle with complex layouts and often yield inaccurate answers. To address these limitations, we present ST-Raptor, an agentic system for semi-structured table QA. ST-Raptor offers an interactive analysis environment that combines visual editing, tree-based structural modeling, and agent-driven query resolution to support accurate and user-friendly table understanding. Experimental results on both benchmark and real-world datasets demonstrate that ST-Raptor outperforms existing methods in both accuracy and usability. The code is available at \emph{\blue{\url{https://github.com/weAIDB/ST-Raptor}}}. And a demonstration video can be found on YouTube\footnote{\emph{\blue{\url{https://youtu.be/9GDR-94Cau4}}}}.
\end{abstract}




\maketitle

\section{INTRODUCTION}

Semi-structured tables are pervasive in real-world applications (e.g., Word-based financial reports and Excel spreadsheets for medical records). Although these tables are visually organized in rows and columns, they lack a strict, unified schema. Cells may contain heterogeneous or multimodal content (e.g., checkboxes, nested subtables, or images), be merged across rows or columns, and be governed by multi-level headers that encode implicit hierarchies and semantic groupings. 
These characteristics make semi-structured tables fundamentally different from relational tables and difficult to interpret using traditional data processing tools such as SQL and BI, as well as even advanced vision–language models.

\begin{figure}[!t]
  \centering
  \includegraphics[width=.96\linewidth]{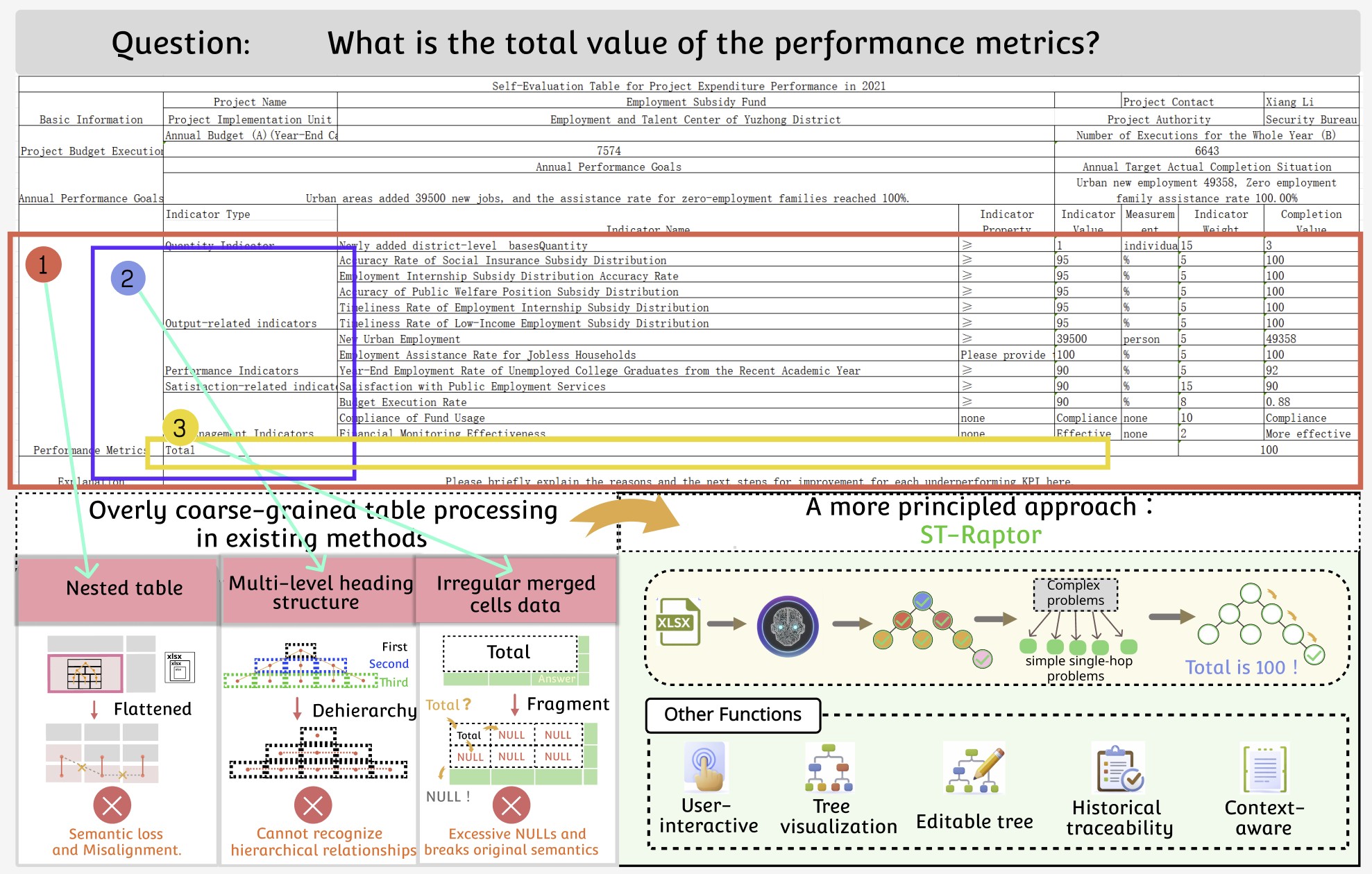}
  \caption{Unique Challenges of Semi-Structured Table QA -- \textnormal{(1) Flattening Nested Tables Causes Semantic Misalignment; (2) Dehierarchizing Multi-Level Headers Obscures Structural Relationships; (3) Fragmenting Merged Cells Breaks Original Semantics.}}
  \vspace{-0.1in}
  \label{fig:domains}
\end{figure}

As shown in Figure~\ref{fig:domains}, existing solutions struggle to handle complex semi-structured tables due to several fundamental limitations: (1) They flatten nested or hierarchically organized tables, causing severe semantic loss and misalignment between headers and content; (2) They decompose multi-level headers into flat token sequences, ignoring the hierarchical relationships critical for interpreting grouped indicators or aggregated statistics; (3) They treat irregular merged cells as sparse matrices filled with artificial \texttt{NULL} values, which breaks the continuity and semantics of logically connected regions (e.g., merged ``Total'' rows). Therefore, downstream question answering (QA) tasks are significantly hindered. For example, answering the total price question in Figure~\ref{fig:domains} requires understanding that the ``Total'' row aggregates across multiple performance metrics with varying weights, which coarse-grained methods often fail to interpret.

\begin{figure*}[!t]
  \centering
  \vspace{-.25em}
  \includegraphics[width=.96\linewidth]{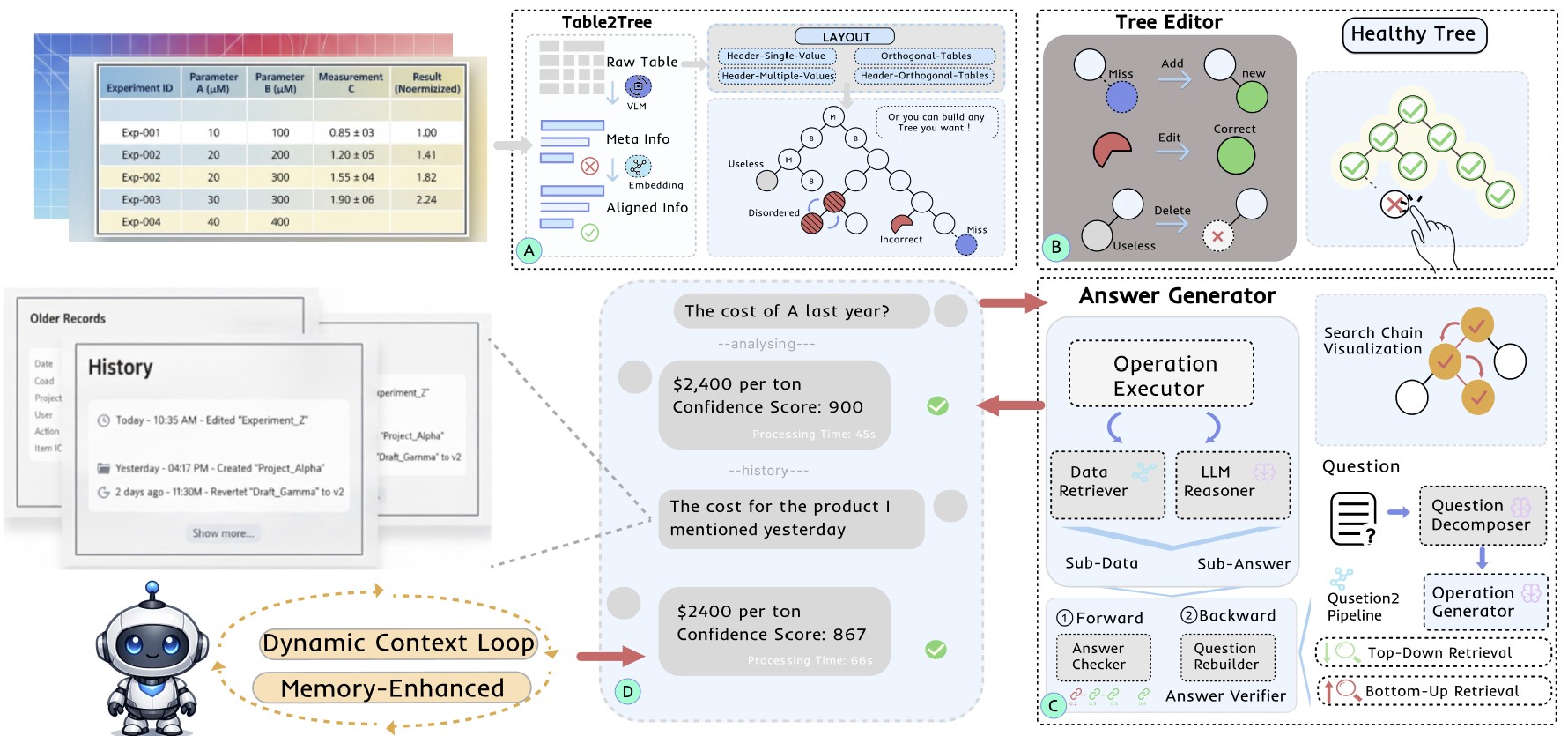}
  \vspace{-0.1in}
  \caption{System Overview of ST-Raptor.}
  \vspace{-0.1in}
  \label{fig:overview}
\end{figure*}

To address these challenges, we present ST-Raptor, a system designed for robust, interactive QA over complex semi-structured tables. At the core of ST-Raptor is the \emph{Hierarchical Orthogonal Tree} (HO-Tree), a layout-aware representation that captures structural relationships among headers, content cells, and merged regions. To construct HO-Trees from raw semi-structured inputs, we adopt a hybrid strategy that integrates rule-based matching with multimodal LLM reasoning: (1) We render semi-structured tables into high-resolution images and prompt a VLM to identify candidate meta-information keys, which are then aligned to table cells via embedding-based similarity matching; (2) We conduct table partitioning guided by the detected meta cells; (3) We then recursively construct HO-Trees based on layout principles such as top-level header identification~\cite{straptor}. Building on the HO-Tree, ST-Raptor defines nine core tree operations that abstract common analytical tasks and enable modular reasoning, offering higher accuracy than a general ``execute-then-reflect'' mechanism. Given user queries over the tables, we automatically decompose them into executable sub-operations, which is guided by a column-type-aware tagging mechanism that distinguishes numerical, categorical, and free-text fields. To ensure answer correctness and faithfulness, we further incorporate a two-stage verification mechanism consisting of forward constraint checks and backward consistency validation.


Compared with our research paper~\cite{straptor}, this demonstration system extends ST-Raptor along three key dimensions:

\noindent\ding{182} \textbf{Interactive and accessible user interface.} To lower the usage barrier, we developed an intuitive and interactive web-based frontend. Users can easily upload spreadsheets, inspect extracted table structures, perform question-answering tasks through a graphical interface. ST-Raptor supports multi-page documents and fragmented table images, covering a broad spectrum of real-world formats, including scanned reports and unstructured Excel sheets.


\noindent\ding{183} \textbf{Editable and user-controllable structure modeling.}  The automatic generation of HO-Trees may occasionally diverge from user expectations due to variations like model mistakes, which limits reliability in practice. To address this, we provide a \emph{Tree Editor} that enables users to visualize, edit, and optimize generated tree structures directly in the frontend. Users can correct misaligned headers, adjust hierarchical groupings, or annotate semantic links using simple drag-and-drop operations. Moreover, the editor supports manual construction of HO-Trees from scratch, enabling the system to generalize beyond semi-structured table QA.


\afterpage{
    \begin{figure*}[t] 
        \centering
        \includegraphics[width=1.0\textwidth,height=0.21\textheight]{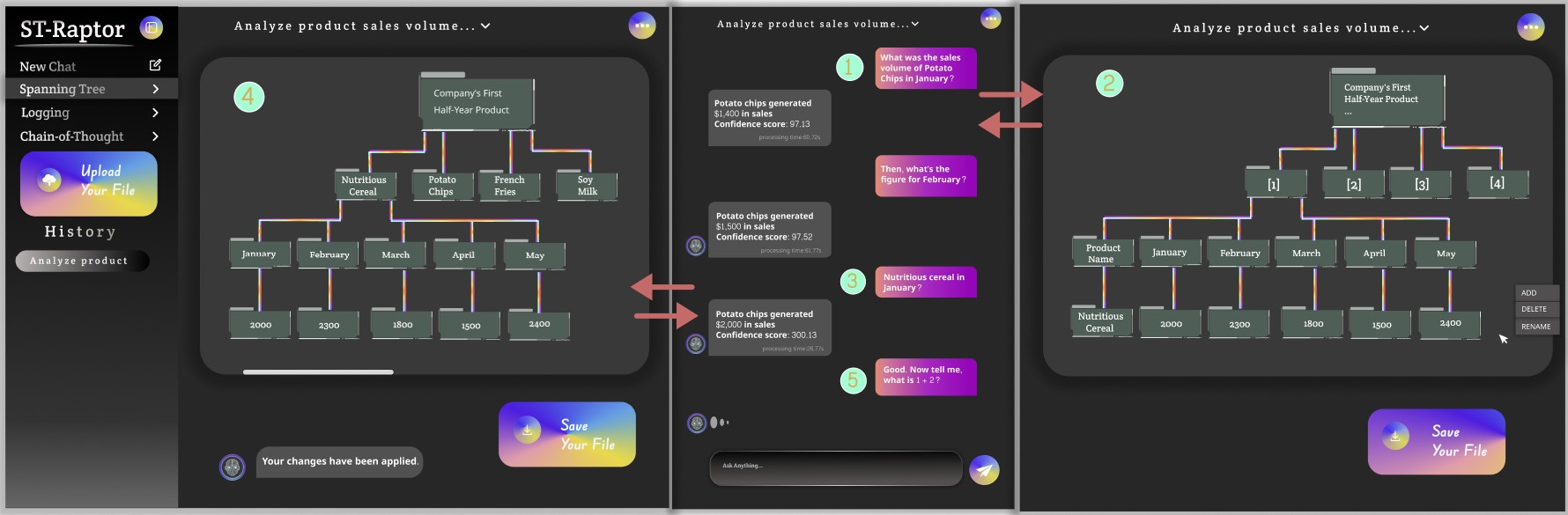}
        \vspace{-.5em}
        \caption{Snapshot of ST-Raptor.}
        \label{fig:frontend}
       \vspace{-.5em}        
    \end{figure*}
}
\noindent\ding{184} \textbf{Agent-driven orchestration for multi-turn and multimodal interactions.} Real-world usage often involves multi-turn interactions, where follow-up questions depend on implicit context. We therefore design an \emph{Orchestration Agent Module} that acts as an AI gateway between users and underlying LLMs. The Agent leverages context-aware attention to resolve ambiguous references and reconstruct complete queries. When processing uploaded images, it invokes specialized vision-language models (VLMs) for table extraction. In multimodal and multi-file settings, it performs precise table localization, routes queries to appropriate processing modules (e.g., retrieval or aggregation), and delivers coherent final answers.



The remainder of this paper is organized as follows. Section~\ref{sec:overview} presents the ST-Raptor system architecture and key components, followed by the HO-Tree construction and interactive Tree Editor. Section~\ref{sec:demo} demonstrates representative real-world usage scenarios, and Figure~\ref{fig:evaluation} reports quantitative evaluation results.

\section{ST-RAPTOR OVERVIEW}
\label{sec:overview}
As discussed, nested regions, multi-level headers, and merged cells in semi-structured tables violate the assumptions of traditional parsers and hinder downstream reasoning. To address these, ST-Raptor transforms complex tables into editable, semantically aligned tree structures for robust and context-aware QA (Figure~\ref{fig:overview}).

\hi{Table2Tree.} The \texttt{Table2Tree} (\bfit{A}) supports multimodal ingestion from raw table files (e.g., \texttt{.xlsx}, \texttt{.csv}, \texttt{.html}, \texttt{.md}) and screenshots. Tables are first rendered via a headless browser to generate high-resolution images for VLMs, which extract candidate meta-information cells as JSON-like key-value structures. These are further validated using embedding-based similarity metrics to filter true header or label cells. Based on table titles and layout semantics, the module then recursively constructs the layout-aware HO-Tree, linking content cells to their semantic headers while preserving hierarchical relationships. When multiple interrelated sheets exist, ST-Raptor unifies them under a single root for global reasoning

\hi{Tree Editor.} Automatically constructed HO-Trees can suffer from noisy predictions or domain-specific inconsistencies. The \texttt{Tree Editor} (\bfit{B}) enables users to refine the tree structure via an intuitive drag-and-drop web interface. Common operations include node reordering, type correction, or subtree pruning. Edits are encoded in JSON and recursively parsed to update the backend representation. This interactive layer is particularly useful in industrial settings where tables are large and slight changes (e.g., removing one employee’s row) should not trigger costly re-parsing of the entire table. Users may also construct HO-Trees from scratch for custom hierarchical modeling needs.

\hi{Answer Generator.} The \texttt{Answer Generator} (\bfit{C}) supports complex, multi-turn QA by decomposing input questions into sub-operations aligned with the HO-Tree structure. Using a pipeline approach, the system performs both top-down and bottom-up subtree retrieval, enabling precise execution over deeply nested layouts. To enhance reliability, ST-Raptor introduces a dual-phase verification scheme: forward verification checks the logic and execution trace of the sub-operations, while backward verification rephrases the question and ensures answer consistency. For instance, given a query like ``What is the average completion rate across all KPI categories?'', ST-Raptor decomposes it into sub-steps (e.g., locate KPI subtree $\rightarrow$ extract completion rates $\rightarrow$ compute average), thereby reducing hallucination risks.



\hi{Orchestration Agent.} Finally, real-world usage often involves multi-turn interactions where follow-up queries depend on prior context (e.g., ``this product'' or ``the table from yesterday''). The \texttt{Orchestration Agent} (\bfit{D}) serves as the controller that tracks query history, resolves ambiguous references, and routes user queries to the appropriate modules. It is equipped with a dynamic memory loop that supports seamless switching across files, tables, and previous turns, thereby enabling fluid and precise user-model interaction. For example, if a user first asks about ``Product A'' and then asks ``What is the profit of this product?'' the agent resolves ``this product'' to ``Product A'' by referring to previous context.

\afterpage{
    \begin{figure*}[t] 
        \centering
        \vspace{-.1em}
        \includegraphics[width=.98\textwidth]{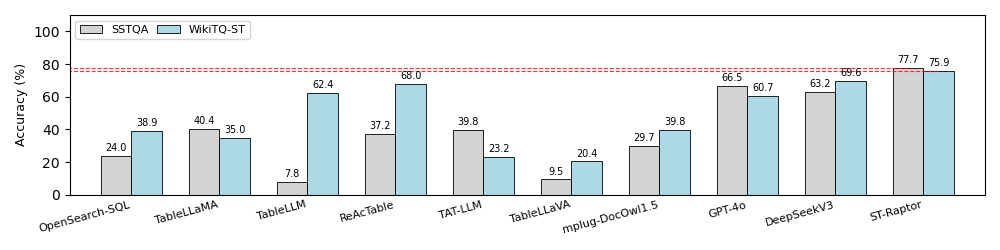}
        \caption{Performance Comparison on the SSTQA and WikiTQ Datasets.}
        \label{fig:evaluation}
    \end{figure*}
}

\section{DEMONSTRATION}
\label{sec:demo}

\subsection{End-to-End Experience}

This section presents the ST-Raptor system interface. As shown in Figure~\ref{fig:frontend}, users can submit tables to generate HO-Trees and conduct multi-turn question-answering. The specific steps are as follows:

\noindent\underline{\textit{(1) Uploading Tables.}} Users need to configure the LLM, VLM, and text embedding model on the model configuration page, and upload multiple tables to be analyzed in the text box. After clicking upload, the model will begin constructing the HO-Trees. Upon successful generation, a notification stating ``HO-Tree generated successfully'' will appear in the upper-right corner of the page, and the resulting files will be saved.

\noindent\underline{\textit{(2) Tree Visualization and Editing.}} Once the HO-Tree is generated, users can view and modify the tree structure through the following operations, which help users better understand the structure of the HO-Tree and generate more accurate answers: (i) Click a node to expand the child nodes connected to it; click again to collapse the subtree. (ii) Right-click to view the menu bar to rename, delete, or create new child nodes. (iii) Hover the mouse over a node to view specific node content in the upper right corner. (iv) Click save to synchronize the constructed tree structure to the LLM for question-answering. If users need to build from scratch, we also support related operations; they can create a root node by clicking the plus sign in the middle of the blank editor.


\noindent\underline{\textit{(3) Multi-Turn QA.}} Users can fill in the questions they need to ask in the input box. After submission, the Agent will complete the subject of the question and feed it into the LLM. After the LLM generates the answer and confidence score, the Agent optimizes the output into a more understandable form based on the question, and counts the answer generation time to display on the frontend page. The user's historical questions will be listed in the dialog box.


\noindent\underline{\textit{(4) Reasoning Chain Visualization.}} For complex multi-hop queries, on this page, users can view the single-step sub-questions generated by the decomposition of the original complex question. After the question is successfully answered, the page will also display the retrieval path of the HO-Tree. Users can modify the tree structure based on this retrieval path to make it more conducive to searching.


\noindent\underline{\textit{(5) History Session Management.}} Considering that users have the need to backtrack to historical sessions and ask questions, users can select the desired session window to switch sessions, and the Agent will find the corresponding HO-Tree for question-answering.


\noindent\underline{\textit{(6) System Monitoring.}} During system operation, the time for the LLM to generate trees and answers may be long. Users can determine the running status of the system by checking the logs. Furthermore, since users may be interested in the specific processes by which the model generates trees and decomposes questions, we have added detailed log outputs for each process. Users can understand the overall operation process through this module.

\subsection{Scenario 1: Interactive Tree Editing for Query Reorientation}

While ST-Raptor automatically constructs the HO-Tree by leveraging layout cues and model predictions, the resulting tree structure may not always align with user intent due to model uncertainty or layout ambiguity. For instance, in a sales table from the SSTQA benchmark, the system, by default, builds the M-Tree (metadata information in the HO-Tree) using temporal headers (e.g., ``January'', ``February'') and the B-Tree using product names. However, when users are more interested in analyzing the monthly trend of a particular product rather than aggregating data by month, the default structure becomes suboptimal. To address this, ST-Raptor provides a Tree Editor that enables users to interactively adjust the tree layout. With simple drag-and-drop operations, users can restructure the HO-Tree (e.g., promoting product names to the MTree) to reflect their analytical goals. Empirically, such refinements not only improve answer accuracy for downstream queries but also enhance system interpretability and responsiveness.

\subsection{Scenario 2: Complex Query Execution on Real-World Tables}

\begin{sloppypar}
We next evaluate how well ST‑Raptor performs on comparatively complex semi-structured table question answering tasks.\end{sloppypar}

\noindent \textbf{Settings:} For ST-Raptor, we employ Gemini-3.0-preview as the vision-language model, Gemini-2.0 as the general large language model, and text-embedding-v1 as the semantic embedding model. And we compare with advanced baselines, including OpenSearch-SQL~\cite{OpenSearch}, TableLLaMA~\cite{TableLlama}, TableLLM~\cite{TableLLM}, ReAcTable~\cite{ReAcTable}, TATLLM~\cite{TATLLM}, TableLLaVA~\cite{tablellava}, mplug-DocOw1.5~\cite{DocOwl}, and general LLMs like GPT-4o and DeepSeekV3.

\noindent \textbf{Data Collection:}  The 102 tables in SSTQA were meticulously filtered from over 2,031 real-world tables covering 19 typical real-world scenarios (such as administrative and financial management). These tables feature semi-structured characteristics, including nested cells, multi-row/column headers, and irregular layouts, thereby ensuring representativeness in terms of both structure and information. WikiTQ contains a large number of semi-structured tables from Wikipedia, but these tables typically have simple layouts with merged cells and are converted to structured formats during the benchmark preprocessing stage.  We converted 25\% of the above tables into images to match practical usage requirements.

\noindent \textbf{Performance Comparison:}  As shown in Figure~\ref{fig:evaluation}, ST-Raptor achieves the highest accuracy, surpassing the best-performing baseline by 11.2\%. This gain is driven by three key techniques in our demo system: (1) The HO-Tree abstraction decouples layout parsing from logical reasoning and supports intuitive inspection and correction via our visual Tree Editor, enabling users to reshape the table structure before QA. (2) The question decomposition module transforms complex natural language queries into sub-questions aligned with tree operations, boosting the interpretability and reliability of model execution. (3) The Agent coordinates model behavior to perform vision-based extraction on table images and to track context across multi-turn queries, which is essential for handling many error-prone scenarios. Compared to LLM-only or agent-based baselines, ST-Raptor avoids semantic collapse and layout loss by preserving structural hierarchy and enabling human-in-the-loop refinement, leading to superior results not only on SSTQA but also on WikiTQ-ST. That is, even on simpler structured tables (like those in WikiTQ-ST), ST-Raptor maintains an edge due to its hybrid layout-to-logic pipeline and modular, reasoning framework. Instead, Agent-based baselines like ReAcTable and DocOwl struggle with structural comprehension (e.g., flattening multi-level headers and treating each header independently).

\bibliographystyle{ACM-Reference-Format}
\bibliography{sample-base}

\end{document}